\pdfoutput=1

\documentclass[11pt]{article}
\usepackage[]{Template/ACL2023}

\usepackage{times}
\usepackage{latexsym}

\usepackage[T1]{fontenc}

\usepackage[utf8]{inputenc}

\usepackage{microtype}

\usepackage{inconsolata}
\usepackage{pgf-pie}
\usepackage{pgfplots}
\usepackage{tikz}
\usetikzlibrary{positioning, shapes.geometric, arrows, calc, math, shapes}
\pgfplotsset{compat=1.18} 
\usetikzlibrary{patterns}
\usepgfplotslibrary{groupplots}
\usepackage{import}
\usepackage{adjustbox}
\usepackage[disable]{todonotes}
\usepackage{caption}
\usepackage{multirow}
\usepackage{subcaption}
\usepackage{makecell}
\setcellgapes{1pt}
\usepackage{listings}
\usepackage{pifont}
\usepackage{booktabs}
\usepackage[inline]{enumitem}
\usepackage{tikzpeople}
\usepackage{fontawesome}

\usepackage{hyperref}

\usepackage{array}
\usepackage{longtable}
\usepackage{tabularx}
\usepackage{rotating}

\newcommand{\xmark}{\ding{55}}%

\newcommand{\tblu}[1]{\textcolor{mydarkgreen}{ #1 }}
\newcommand{\tred}[1]{\textcolor{mydarkred}{#1}}

\newcommand{\mpara}[1]{\medskip\noindent{\bf #1}}

\newcommand{\commandr}{\texttt{Cmd-R+}}
\newcommand{\gptfour}{\texttt{GPT 4}}

\newcommand{\llamathreeone}{\texttt{Llama 3.1}}

\newcommand{\gemmatwo}{\texttt{Gemma 2}}
\newcommand{\qwen}{\texttt{Qwen 2.5}}
\newcommand{\jamba}{\texttt{Jamba 1.5}}

\definecolor{boxcolor}{HTML}{1E88E5} %
\definecolor{textcolor}{HTML}{FFFFFF} %
\definecolor{arrowcolor}{HTML}{FF5722} %
\definecolor{highlight}{HTML}{FFC107} %

\tikzset{%
    myarrow/.style={-Stealth, thick},
    process/.style={rectangle, draw, thick, align=center, minimum height=0.7cm, minimum width=1.2cm, font=\footnotesize},
    data/.style={cylinder, shape border rotate=90, draw, thick, minimum height=0.7cm,
    minimum width=1cm, align=center, aspect=0.3},
}%

\colorlet{mypink}{blue!20!red!30!white}
\colorlet{mygreen}{green!60!blue!40!white}
\colorlet{myred}{blue!10!red!50}
\colorlet{mydarkred}{blue!20!red!80}
\colorlet{mydarkestred}{blue!30!red!70!black}
\colorlet{mydarkpink}{red!60!blue!60}
\colorlet{mydarkestgreen}{green!60!blue!80!black}
\colorlet{mydarkgreen}{green!30!blue!90}
\colorlet{myblue}{red!10!blue!30!white}
\colorlet{mydarkblue}{red!20!blue!70!white}
\colorlet{mylightgrey}{black!5}

\title{
\raisebox{-0.25\height}
{\includegraphics[width=0.8cm]{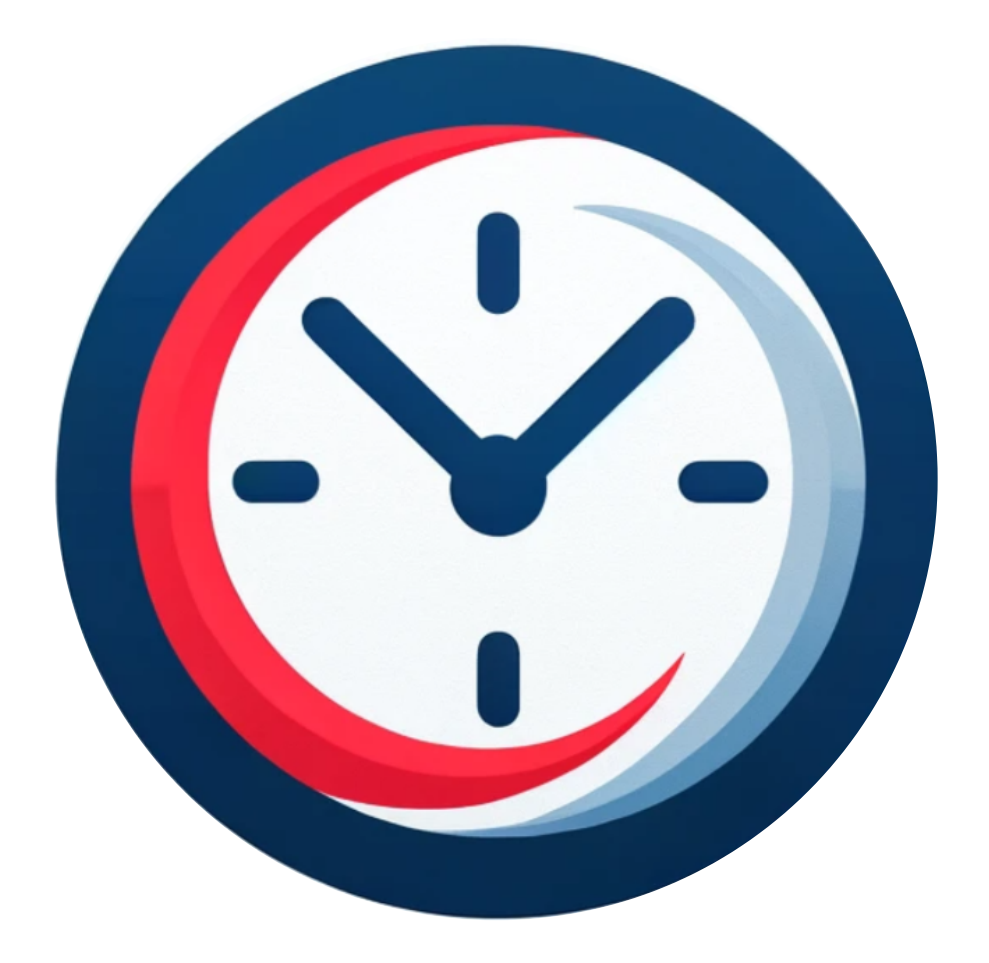}
}
A Study into Investigating Temporal Robustness of LLMs}

\author{Jonas Walat \\
  L3S Research Center \\
  Hannover, Germany \\
  \texttt{jonas.wallat@l3s.de} \\\And
  Abdelrahman Abdallah \\  {\bf Adam Jatowt} \\
  University of Innsbruck \\
  Innsbruck, Austria \\
  \texttt{<first.last>@uibk.ac.at} \\
  \And
  Avishek Anand \\
  TU Delft \\
  Deflt, The Netherlands \\
  \texttt{avishek.anand@tudelft.nl}} 

\begin{document}
\maketitle
\begin{abstract}
Large Language Models (LLMs) encapsulate a surprising amount of factual world knowledge. 
However, their performance on \textit{temporal questions} and historical knowledge is limited because they often cannot understand temporal scope and orientation or neglect the temporal aspect altogether.
In this study, we aim to measure precisely how robust LLMs are for question answering based on their ability to process temporal information and perform tasks requiring temporal reasoning and temporal factual knowledge. 
Specifically, we design eight time-sensitive
robustness tests for factual information to check the sensitivity of six popular LLMs in the zero-shot setting.
Overall, we find LLMs lacking temporal robustness, especially to temporal reformulations and the use of different granularities of temporal references. 
We show how a selection of these eight tests can be used automatically to judge a model’s temporal robustness for user questions on the fly. Finally, we apply the findings of this study to improve the temporal QA performance by up to 55\%.
\end{abstract}

\section{Introduction}
\label{sec:intro}

Despite the strong zero- and few-shot performance of LLMs, it has been recently pointed out that LLMs suffer from a partial or imprecise understanding of the \textit{temporal scope, orientation, and reasoning} expressed in text~\cite{chan:2023:arxiv:chatgpttemporalcausaldisscourse,Yuan:2023:BioNLP,wallat2024temporal,jain-etal-2023-language-models}.
The inaccurate understanding of temporal orientation and grounding raises concerns regarding the effectiveness of LLMs over a range of tasks involving temporal reasoning and intents like question-answering and search over historical sources~\cite{DBLP:conf/sigir/WangJY22}, QA over legal and personal temporal collections~\cite{qin:2020:www:ltrpersonalsearch, zamani:2017:www:sitcontextpersonalsearch, gupta:2019:www:personalizedspellcorrectionPersonalSearch}, or fact checking~\cite{lee:2020:arxiv:lmsfactcheckers, nakov:2021:ijcai:automaticFactChecking}.
Moreover, questions with temporal aspects are relatively rare in many current QA datasets and may thus go undetected in offline evaluations.
In this paper, we study the ability of LLMs~\cite{Brown2020GPT3} in temporal QA tasks, given their excellent ability of language understanding and reasoning.

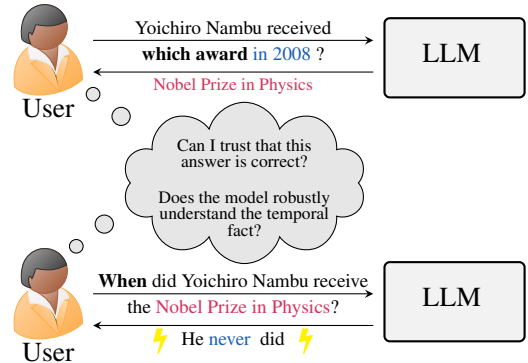
\begin{figure}[t]
    \centering
        {\begin{tikzpicture}[scale=0.6, transform shape]
    \tikzmath{
    \boxwidth = 0.027\textwidth;
    \boxheight = 7;
    \spacing = 0.2;
    }
     \draw[local bounding box=rect0, rounded corners=3pt, draw=white]  (0,0) rectangle ($(0,0) + (\boxwidth,\boxheight)$);
        \begin{scope}[
            x={($(rect0.south east)-(rect0.south west)$)},
            y={($(rect0.north west)-(rect0.south west)$)},
            shift={(rect0.south west)}
        ]

            \node[alice,minimum size=1.5cm] (a1) at (0.1,0.87) {\LARGE{User}};

            \draw[fill=black!5, rounded corners=2pt, thick] (0.7,0.72) rectangle (0.95,0.97);
            \node (llm1) at (0.82, 0.86) {\LARGE{LLM}};

            \draw[-stealth] (0.18, 0.9) -- (0.68, 0.9) node[midway, above] {\large Yoichiro Nambu received} node[midway, below] {\large \textbf{which award} \tblu{in 2008}?};
            
            \draw[stealth-] (0.18, 0.8) -- (0.68, 0.8) node[midway, below] {\tred{Nobel Prize in Physics}};

            \node[alice,minimum size=1.5cm] (a2) at (0.1,0.1) {\LARGE{User}};
            \node[cloud, draw, fill=gray!20, aspect=2] at (0.45, 0.45) {\large \textcolor{gray!20}{the answer correct?}}; 
            \draw[fill=gray!20] (0.18,0.73) circle [radius=0.15cm];
            \draw[fill=gray!20] (0.23,0.66) circle [radius=0.2cm];
            \draw[fill=gray!20] (0.15,0.25) circle [radius=0.15cm];
            \draw[fill=gray!20] (0.2,0.32) circle [radius=0.2cm];
            \node at (0.45,0.57) {Can I trust that this};
            \node at (0.45,0.52) {answer is correct?};
            \node at (0.45,0.4) {Does the model robustly};
            \node at (0.45,0.35) {understand the temporal};
            \node at (0.45,0.3) {fact?};

            \draw[fill=black!5, rounded corners=2pt, thick] (0.7,-0.05) rectangle (0.95,0.2);
            \node (llm2) at (0.82, 0.09) {\LARGE{LLM}};

            \draw[-stealth] (0.18, 0.1) -- (0.68, 0.1) node[midway, sloped, above] {\large \textbf{When} did Yoichiro Nambu receive} node[midway, below] {\large the \tred{Nobel Prize in Physics}?};

            \draw[stealth-] (0.18, 0.0) -- (0.68, 0.0) node[midway, below] {\large He \tblu{never} did};
            \node at (0.30, -0.05) {\LARGE \color{yellow}{\faBolt}};
            \node at (0.56, -0.05) {\LARGE \color{yellow}{\faBolt}};
           
        \end{scope}

\end{tikzpicture}}
    \setlength{\abovecaptionskip}{3pt}
    \caption{We investigate the robustness of temporal understanding with a set of tests (here: temporal reversal). By asking the inverse question and looking for consistency between the two answers, we can study if the model understands the temporal-factual information.}
    \label{fig:fig0}
    \vspace*{-4mm}
\end{figure}

\begin{figure*}[ht]
    \centering
        {\begin{tikzpicture}[scale=0.6, transform shape]
    \tikzmath{
    \boxwidth = 4.5;
    \boxheight = 4;
    \spacing = 0.2;
    }
     \draw[local bounding box=rect0, rounded corners=3pt, draw=mydarkgreen,thick]  (0,0) rectangle ($(0,0) + (\boxwidth,\boxheight)$);
        \begin{scope}[
            x={($(rect0.south east)-(rect0.south west)$)},
            y={($(rect0.north west)-(rect0.south west)$)},
            shift={(rect0.south west)}
        ]
            \draw[rounded corners=3pt,
            draw=mydarkgreen,
            fill=mydarkgreen] (0, 0.9) rectangle (1.0,1.0);
            \node at (0.5, 0.95) {\textbf{\color{white}{Temporal Reversal}}};

            \node at (0.5, 0.8) {Yoichiro Nambu received};
            \node at (0.5, 0.7) {which award \tblu{in 2008}?};
            \node (q1) at (0.5, 0.6) {[\tred{Nobel Prize in Physics}]};

            \node (q2) at (0.5, 0.30) {When did Yoichiro Nambu};
            \node at (0.5, 0.20) {receive the \tred{Nobel Prize in}};
            \node at (0.5, 0.10) {\tred{Physics}? [\tblu{2008}]};

            \draw[-stealth] (q1.south) -- (q2.north);
           
        \end{scope}
    
    \draw[local bounding box=rect, rounded corners=3pt, draw=mydarkgreen,thick]  ($($(rect0.south east)$) + (\spacing,0)$) rectangle
    ($($($(rect0.south east)$) + (\spacing,0)$) + ($(\boxwidth,\boxheight)$)$);

        \begin{scope}[
            x={($(rect.south east)-(rect.south west)$)},
            y={($(rect.north west)-(rect.south west)$)},
            shift={(rect.south west)}
        ]
            \draw[rounded corners=3pt,
            draw=mydarkgreen,
            fill=mydarkgreen] (0, 0.9) rectangle (1.0,1.0);
            \node at (0.5, 0.95) {\textbf{\color{white}{Event Dating}}};

            \node at (0.5, 0.8) {Event:};
            \node at (0.5, 0.7) {The fifth ICC T20 World};
            \node (q1) at (0.5, 0.6) {Cup in cricket is held in};
            \node at (0.5, 0.5) {Bangladesh};

            \node (q2) at (0.5, 0.3) {Predict date:};
            \node at (0.5, 0.2) {[\tblu{16-03-2014}]};

        \end{scope}
    
    \draw[local bounding box=rect2, rounded corners=3pt, draw=mydarkgreen,thick]  (0,-\spacing)  rectangle    ($(0,0) + (\boxwidth,-\boxheight-\spacing)$);
        \begin{scope}[
            x={($(rect2.south east)-(rect2.south west)$)},
            y={($(rect2.north west)-(rect2.south west)$)},
            shift={(rect2.south west)}
        ]
            \draw[rounded corners=3pt,
            draw=mydarkgreen,
            fill=mydarkgreen] (0, 0.9) rectangle (1.0,1.0);
            \node at (0.5, 0.95) {\textbf{\color{white}{Event Ordering}}};

            \node at (0.5, 0.8) {\underline{Event A}: Albania and};
            \node at (0.5, 0.7) {Croatia join NATO};
            \node (q1) at (0.5, 0.6) {\underline{Event B}: Ecuador declares};
            \node at (0.5, 0.5) {independence from Spain};

            \node (q2) at (0.5, 0.3) {Predict order:};
            \node at (0.5, 0.2) {A \tblu{before} B? [false]};

        \end{scope}
        
    \draw[local bounding box=rect3, rounded corners=3pt, draw=mydarkgreen,thick]  ($($(rect2.south east)$) + (\spacing,0)$)  rectangle ($($($(rect2.south east)$) + (\spacing,0)$) + ($(\boxwidth,\boxheight)$)$);
        \begin{scope}[
            x={($(rect3.south east)-(rect3.south west)$)},
            y={($(rect3.north west)-(rect3.south west)$)},
            shift={(rect3.south west)}
        ]
            \draw[rounded corners=3pt,
            draw=mydarkgreen,
            fill=mydarkgreen] (0, 0.9) rectangle (1.0,1.0);
            \node at (0.5, 0.95) {\textbf{\color{white}{Temporal Removal}}};

            \node at (0.5, 0.8) {What prize did Hemingway};
            \node (q1) at (0.5, 0.7) {win \underline{\tblu{in 1954}}?};

            \node (q2) at (0.5, 0.3) {What prize did Hemingway};
            \node at (0.5, 0.2) {win \_\_\_?};

            \draw[-stealth] (q1.south) -- (q2.north);
            
        \end{scope}

    \draw[local bounding box=rect4, rounded corners=3pt, draw=mydarkgreen,thick]  ($($(rect3.south east)$) + (\spacing,0)$)  rectangle ($($($(rect3.south east)$) + (\spacing,0)$) + ($(\boxwidth,\boxheight)$)$);
        \begin{scope}[
            x={($(rect4.south east)-(rect4.south west)$)},
            y={($(rect4.north west)-(rect4.south west)$)},
            shift={(rect4.south west)}
        ]
            \draw[rounded corners=3pt,
            draw=mydarkgreen,
            fill=mydarkgreen] (0, 0.9) rectangle (1.0,1.0);
            \node at (0.5, 0.95) {\textbf{\color{white}{Temporal Positioning}}};

            \node at (0.5, 0.8) {What prize did Hemingway};
            \node (q1) at (0.5, 0.7) {win \underline{\tblu{in 1954}}?};

            \node (q2) at (0.5, 0.3) {\underline{\tblu{In 1954}}, what prize};
            \node at (0.5, 0.2) {did Hemingway win?};

            \draw[-stealth] (q1.south) -- (q2.north);

        \end{scope}

    \draw[local bounding box=rect6, rounded corners=3pt, draw=mydarkgreen,thick]  ($($(rect.south east)$) + (\spacing,0)$) rectangle
    ($($($(rect.south east)$) + (\spacing,0)$) + ($(\boxwidth,\boxheight)$)$);

        \begin{scope}[
            x={($(rect6.south east)-(rect6.south west)$)},
            y={($(rect6.north west)-(rect6.south west)$)},
            shift={(rect6.south west)}
        ]
            \draw[rounded corners=3pt,
            draw=mydarkgreen,
            fill=mydarkgreen] (0, 0.9) rectangle (1.0,1.0);
            \node at (0.5, 0.95) {\textbf{\color{white}{Relativization}}};

            \node at (0.5, 0.8) {What prize did Hemingway};
            \node (q1) at (0.5, 0.7) {win \underline{\tblu{in 1954}}?};

            \node (q2) at (0.5, 0.3) {What prize did Hemingway};
            \node at (0.5, 0.2) {win \underline{\tblu{69 years ago}}?};

            \draw[-stealth] (q1.south) -- (q2.north);

        \end{scope}

    \draw[local bounding box=rect5, rounded corners=3pt, draw=mydarkgreen,thick]  ($($(rect4.south east)$) + (\spacing,0)$)  rectangle ($($($(rect4.south east)$) + (\spacing,0)$) + ($(\boxwidth,\boxheight)$)$);
        \begin{scope}[
            x={($(rect5.south east)-(rect5.south west)$)},
            y={($(rect5.north west)-(rect5.south west)$)},
            shift={(rect5.south west)}
        ]
            \draw[rounded corners=3pt,
            draw=mydarkgreen,
            fill=mydarkgreen] (0, 0.9) rectangle (1.0,1.0);
            \node at (0.5, 0.95) {\textbf{\color{white}{Fact Checking}}};

            \node at (0.5, 0.8) {\underline{Fact}: Sri Lanka};
            \node at (0.5, 0.7) {imposes a new levy for};
            \node (q1) at (0.5, 0.6) {those leaving the country};
            \node at (0.5, 0.5) {\tblu{in 2023}};

            \node (q2) at (0.5, 0.3) {Classify fact:};
            \node at (0.5, 0.2) {[true/false/conflicting]};

        \end{scope}

    \draw[local bounding box=rect7, rounded corners=3pt, draw=mydarkgreen,thick]  ($($(rect6.south east)$) + (\spacing,0)$) rectangle
    ($($($(rect6.south east)$) + (\spacing,0)$) + ($(\boxwidth,\boxheight)$)$);

        \begin{scope}[
            x={($(rect7.south east)-(rect7.south west)$)},
            y={($(rect7.north west)-(rect7.south west)$)},
            shift={(rect7.south west)}
        ]
            \draw[rounded corners=3pt,
            draw=mydarkgreen,
            fill=mydarkgreen] (0, 0.9) rectangle (1.0,1.0);
            \node at (0.5, 0.95) {\textbf{\color{white}{Year Shift}}};

            \node at (0.5, 0.8) {What prize did Hemingway};
            \node at (0.5, 0.7) {win \underline{\tblu{in 1954}}?};
            \node (q1) at (0.5, 0.6) {};

            \node (q2) at (0.5, 0.3) {What prize did Hemingway};
            \node at (0.5, 0.2) {win in [\tblu{1953/1949/1944}]?};

            \draw[-stealth] (q1.south) -- (q2.north);

        \end{scope}

    \draw[local bounding box=rect8, rounded corners=3pt, draw=white,thick]
    ($($(rect5.south east)$) + (8*\spacing,0)$) rectangle
    ($($($(rect7.south east)$) + (8*\spacing,0)$) + ($(1.35*\boxwidth,\boxheight)$)$);

        \begin{scope}[
            x={($(rect8.south east)-(rect8.south west)$)},
            y={($(rect8.north west)-(rect8.south west)$)},
            shift={(rect8.south west)}
        ]
            \draw[rounded corners=3pt,
            draw=mydarkred,
            fill=mydarkred] (0, 0.95) rectangle (1.0,1.0);
            \node at (0.5, 0.975) {\textbf{\color{white}{Applications}}};

            \draw[rounded corners=3pt,
            draw=mydarkred, thick] (0, 0.65) rectangle (1.0, 0.94);
            \node at (0.5, 0.9) {\textbf{Understanding temporal robustness}};
            \node at (0.5, 0.83) {Section \ref{sec:results}};
            \node at (0.5, 0.75) {Temporal robustness findings:};
            \node at (0.5, 0.7) {Phrasing matters, ordering is hard, ...};

            \draw[rounded corners=3pt,
            draw=mydarkred, thick] (0, 0.35) rectangle (1.0, 0.64);
            \node at (0.5, 0.6) {\textbf{Refining trust /w automatic tests}};
            \node at (0.5, 0.53) {Section \ref{sec:automatic_tests}};
            \node at (0.5, 0.45) {Can one trust the output when no};
            \node at (0.5, 0.4) {ground-truth is available?};

            \draw[rounded corners=3pt,
            draw=mydarkred, thick] (0, 0.0) rectangle (1.0, 0.34);
            \node at (0.5, 0.3) {\textbf{Reformulations for better QA perf.}};
            \node at (0.5, 0.23) {Section \ref{sec:findings_application}};
            \node at (0.5, 0.12) {Robustness findings are useful:};
            \node at (0.5, 0.07) {Guidelines on how to phrase questions};

        \end{scope}

    \draw[-latex, line width=3.5pt] (18.85,3) -- (20,3);
    \draw[-latex, line width=3.5pt] (18.85,-3) -- (20,-3);
    \draw[-latex, line width=3.5pt] (18.85,0) -- (20,0);

\end{tikzpicture}}
    \setlength{\abovecaptionskip}{2pt}
    \caption{Overview of the different tests in our robustness test suite for temporal factual QA. We suggest a suite of several tests that are useful in multiple applications: 1) helping to assess the temporal robustness of LLMs for temporal QA, 2) Calibrating user trust at inference time, and 3) as guidelines on how to reformulate arbitrary (temporal) questions to improve QA performance.}
    \label{fig:overview_tests}
    \vspace*{-4mm}
\end{figure*}
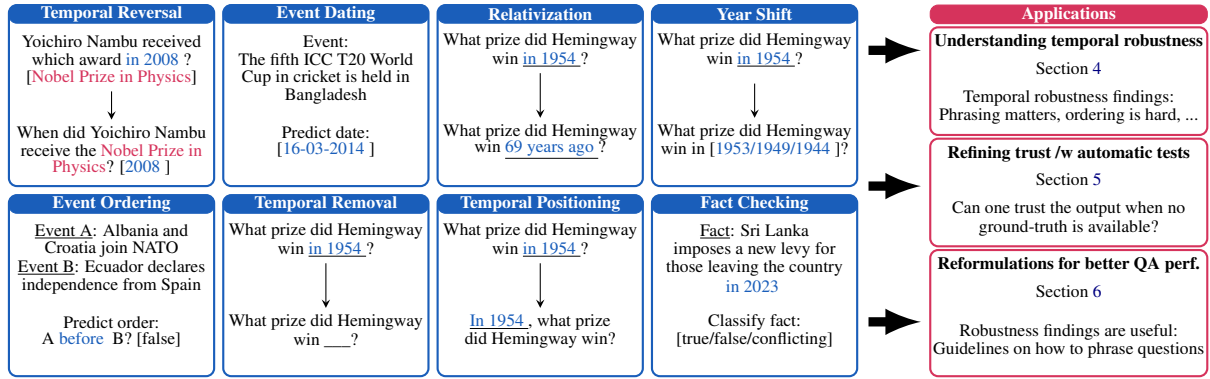

Consider the following factual question: ``Who was the prime minister of Pakistan in 1992?''. The answer to this question is \emph{Nawaz Sharif}. 
When asked this question to an LLM (say a Mistral-7B model) by just changing the year \textit{1992} to \textit{1995}, \textit{2010}, or \textit{1970} -- we still observe the same response.
This simple test indicates a disregard for the time information, possibly due to popularity bias.
In this paper, we propose a series of tests that help identify when LLMs can fail due to improper understanding of time and handling of temporal information in question-answering (Figure~\ref{fig:fig0}).
Unlike earlier literature that focuses on characterizing temporal failures~\cite{wallat2024temporal}, we provide concrete test cases and question reformulations to automatically determine the sensitivity of LLMs to temporal information in questions or lack thereof. The tests evaluate both the LLMs' storage abilities and robust temporal reasoning for correct question understanding.
This set of tests can be used as LLMs' pre-deployment tests in combination with the regular task performance.

In this paper, we first introduce a range of automatic transformations that manipulate temporal information in questions to estimate the robustness of LLMs towards temporal questions. 
Our transformations consider multiple and diverse interventions of the time component in questions like \textit{temporal removal, positioning, year-shift}, and \textit{temporal reversal} (cf. Figure~\ref{fig:overview_tests}). 
Our tests are grounded on well-understood properties and challenges of event-based question answering and temporal information retrieval like temporal ordering and dating~\cite{campos2015survey,DBLP:journals/ftir/KanhabuaBN15,setty2017modeling}. 
Secondly, we perform extensive benchmarking over three well-known temporal QA datasets~\cite{chen:2021:NeurIPS:timesensitiveqa,wang2021improving,DBLP:conf/sigir/WangJY22}, two additional test sets composed of sentences containing a series of temporal facts and important historical events, and six popular LLMs~\cite{Brown2020GPT3}. 
Our results show that although LLMs can answer the questions in their original form, they struggle under certain temporal transformations -- lacking a robust understanding of the underlying temporal fact.
Specifically, we see performance drops of 47-67\% under the temporal reversal query transformation.
Although LLMs can successfully date many events on the year granularity, they find it hard to date them on a day granularity (performance drops by 27-75\%). 
Also, and surprisingly, although LLMs can perform event dating reasonably well, they find it difficult to order events chronologically. 
Additionally, the findings of this study can be applied and be valuable beyond informativeness:
We showcase that a subset of the tests can be used automatically to understand better whether the model's answer is correct without access to the ground-truth answers. Lastly, we show that by applying the findings from this study to new temporal questions, we can use guided reformulations to improve the relative QA performance by up to 55\%. 
Our proposal of benchmarking allows for more precise gauging of how robust LLMs are when it comes to the temporal knowledge and abilities they possess. The findings are applicable in multiple scenarios, and code \& data are available\footnote{\url{https://github.com/jwallat/temporalrobustness}}.

\section{Related Work}
\label{subsec:bio}

\begin{table*}[ht!]
\scriptsize
\begin{tabular}{lllccc}
\toprule
\textbf{Dataset} & \textbf{Example Question} & \textbf{Answer} & \textbf{\#Qs} & \textbf{Scope} & \textbf{Description}\\ \hline
\textbf{ArchivalQA }       & \texttt{What was Ankara's official aid bill for in 1997?} & Cyprus               & 7,500 &1987-2007 & detailed quest. \\ 
\textbf{Wikidata} & \texttt{Yoichiro Nambu received which award in 2008?}& Nobel Prize& 10,000& 1907-2018&people\\
\textbf{Temporal Claims} & \texttt{China reports military clash in Henan province in 2022} & False& 4,196& 1000-2023&claims\\
\textbf{Wikipedia Events} & \texttt{Former Pope Benedict XVI dies at the age of 95.}& Dec. 31 2022& 23,550& 1750-2023&events\\
\bottomrule
\end{tabular}
\caption{Overview of the temporal source datasets used to create the robustness tests in this study.}
\label{tab:datasets_overview}
\vspace*{-2mm}
\end{table*}

\subsection{Time-aware Pre-trained Language Models}
The pretraining approaches used in Language Models (e.g., BERT \cite{BERT}) do not specifically consider or model temporal information.
Several time-focused enhancements and adaptations of language models have been then proposed recently. A naive approach relies on training different versions of a language model on time-segmented portions of data \cite{HistBert}. This results in multiple language models that require an alignment stage as postprocessing. Other solutions explore dynamic word embeddings \cite{DynamicWordEmbedd}.

More advanced approaches incorporate temporal knowledge during the pretraining stage \cite{giulianelli2020analysing, Dhingra_2022,  rosin2022time, Wang2023}. A simple yet effective modification to pre-training is proposed in \cite{Dhingra_2022}, where the masked language modeling (MLM) objective is parameterized with timestamp information. %
\citet{rosin2022temporal} propose to enhance the self-attention mechanism by integrating timestamp information for updating attention scores. A time-aware prompting strategy for text generation has been proposed by \cite{cao2022timeawarepromptingtextgeneration}.
In a more recent study, \citet{cole2023salient} integrate content time into the transformer encoder-decoder architecture (T5 model). They mask time expressions in content and conduct experiments on different temporal tasks. \citet{Wang2023} utilize transformer encoders to leverage both the timestamp and content time (i.e., temporal expressions) in three novel pre-training tasks: document timestamping, temporal expression masking, and temporal information swapping.
Relatedly, \citet{han:2021:ACL:ECONET} train models by masking events and temporal indicators and \citet{yang:2023:EMNLP:OnceUponATime} introduce a novel timeline reconstruction task.

\subsection{Benchmarking LLMs Robustness}
\label{sec:rw_benchmarks}

Several studies have examined the temporal reasoning capabilities of LLMs \cite{wang2023tram,xiong2024large,chu2023timebench,jain-etal-2023-language-models,zhou-etal-2019-going,fatemi2024testtimebenchmarkevaluating}. 
\citet{yuan2023back} studied explanatory capabilities of LLMs when forecasting future events, while \citet{chan:2023:arxiv:chatgpttemporalcausaldisscourse} analyzed Chat-GPT on inter-sentential relations including temporal and causal relations. \citet{chu2023timebench} introduced a hierarchical temporal reasoning benchmark called TimeBench and focused on Chain-of-Thought prompting. \citet{wang2023tram} introduced TRAM,
 a temporal reasoning benchmark encompassing temporal aspects of events such as order, arithmetic, frequency, and duration. For interested readers, temporal commonsense reasoning datasets and approaches have been overviewed in \cite{wenzel2023}.

Temporal factual knowledge has also been the focus of several recent QA datasets created to analyze the performance of LLMs~\cite{chen:2021:NeurIPS:timesensitiveqa,DBLP:conf/sigir/WangJY22,Dhingra_2022,gruber2024,DBLP:conf/www/JiaARSW18,jia2024tiq,mousavi2024dyknowdynamicallyverifyingtimesensitive}. For example, a diagnostic dataset \textit{TempLAMA} introduced in \citet{Dhingra_2022} contains 50k temporally-scoped subject-object relations collected from the snapshot of Wikidata and provided in the cloze-style queries. The authors discuss potential problems related to encoding factual temporal knowledge, such as averaging, forgetting, and poor temporal calibration.
 More recently, \citet{wallat2024temporal} study if LLMs can answer temporal questions and reveal that they struggle with simple perturbations in questions like time relativization or time shift. However, the authors do not introduce a complete test suite for temporal robustness as we do (e.g., event ordering, event dating, fact verification, time positioning, temporal reversal), neither propose automatic question transformations nor demonstrate how temporal QA performance can be improved. \citet{bajpai-etal-2024-temporally} introduce temporally consistent factuality probing and the corresponding dataset constructed from a knowledge graph for measuring the temporal consistency of objects and their relations. In another work, \citet{beniwal-etal-2024-remember} demonstrate that diverse fine-tuning approaches significantly improve the performance of open-source LLMs, reducing errors caused by knowledge gaps.

Our research emphasizes novel approaches for investigating temporal signals, anchoring knowledge in time, and navigating and orienting over timelines utilizing a range of different datasets and models. We propose a set of automatic transformation steps that, given any temporal QA dataset, allow it to be extended to gauge the temporal robustness of LLMs, and we also demonstrate how our approach can enhance QA performance.

\section{Study Details}
\label{sec:study}

We use several data sources to test factual knowledge with time-scoped questions for assessing LLMs' robustness in handling temporal references: WikiData (Time-Sensitive QA~\cite{chen:2021:NeurIPS:timesensitiveqa}), the historical New York Times news archive (ArchivalQA~\cite{DBLP:conf/sigir/WangJY22}), TemporalQuestions~\cite{wang2021improving}, 
major world events from Wikipedia (event dating/ordering), and fact-checked temporal claims crawled from various websites (Temporal Claims~\cite{venktesh:2024:arxiv:quantemp}). 
An overview is given in Table~\ref{tab:datasets_overview}. We elaborate further on the source datasets and model \& implementation details in Appendix~\ref{sec:appendix_implementation_details}.

\mpara{Time Relativization, Removal, Year Shift, Positioning.}
We sample 3k QA pairs from ArchivalQA that end with a year reference (e.g., "in 2019?") and modify the references according to the task (as in \cite{wallat2024temporal}). For relativization, we convert an absolute year reference to a relative one\footnote{Using the answer to "What year do we have?"}. For the year shift, we randomly decide whether we add or deduct $k$ years from the question's original year. Lastly, for the positioning test, we move the time reference from the end of the question (e.g., "... in 2019?") to the front of it ("In 2019, ..."). Examples of these transformations and the remaining tests can be seen in Figure~\ref{fig:overview_tests}.

\mpara{Temporal Reversal.}
We use WikiData information (similar to \citet{saxena:2021:ACL:cronQA}) and interpret these factual statements as quadruples $($$subject$, $relation$, $object$, $time$$)$. In other studies, these quadruples have been used to construct questions such as "Who was the American president in 2012?" Answer: "Obama," asking for the subject or the object of the quadruple. We hypothesize that a thorough, actual understanding of the question's temporal aspect would result in the model being able to answer both the normal (forward) question as well as a reformulation of this question that queries for the time (e.g., "When was Obama president of the USA?" Answer: "2009-2017"). We utilize 10k examples from WikiData for this test since it has quadruples with individual years and the required intervals in which the relation was true. We apply the set of relations used by \citet{saxena:2021:ACL:cronQA} and write templates for the reformulations.   

\mpara{Temporal Fact Checking.}
We use a dataset of manually verified facts crawled from various verification websites \cite{venktesh:2024:arxiv:quantemp}, containing 4,196 temporally scoped claims. Fact verification requires the model to produce a judgment of $true$, $false$, or $contradicting$ for a given claim.

\mpara{Event Dating/Ordering.}
Similar to \cite{DBLP:conf/sigir/WangJY21}, we crawl events from the Wikipedia year pages\footnote{E.g., https://en.wikipedia.org/wiki/2006} to acquire fine-grained dates (containing a day, month, and year) and short descriptions of major events between 1750 and 2023. We then filter out events that contain years in the description, as these would be easy to guess. For the event dating test, we ask the model to reproduce the date for a given event in different granularities: year, month, and date. For the event ordering test, we randomly sample events from the same year or for given distances $k$. We then ask the models to answer which event happened first. The event dating task uses 3k events for each granularity, and the event dating has 3k event pair comparisons.

\begin{table*}[ht]
    \small
    \centering
    \begin{tabular}{lccccccccc}
    \toprule
    \textbf{Model}    &\textbf{Size}& $\leftrightarrow$\textbf{Relativ.}& \textbf{$\downarrow$Removal} & \textbf{Shift}& $\leftrightarrow$\textbf{Reversal}& $\uparrow$\textbf{Facts} & $\leftrightarrow$\textbf{Date}& $\leftrightarrow$\textbf{Order} & $\leftrightarrow$\textbf{Position}\\ \hline
 \llamathreeone{} &8B& -50.0\%& -48.0\%& -70.4\%& -55.6\%& 34.2& -75.4\%& \textbf{-0.2\%}& +4.8\%\\
 \gemmatwo{}&27B& -\textbf{27.9}\%& -46.1\%& -56.9\%& -60.6\%& 42.6& \underline{-42.6\%}& -36.1\%& +3.3\%\\
 \qwen{} &32B& -41.0\%& -48.9\%& -68.2\%& -67.7\%& 31.4& -60.9\%& -67.6\%& \textbf{+0.6\%}\\
 \jamba{} &52B& -29.1\%& -\textbf{39.5}\%& -43.6\%& -62.1\%& \textbf{52.5}& -52.3\%& \underline{-1.0\%}& +3.5\%\\
 \commandr{} &104B& -34.3\%& -\underline{40.3}\%& -54.2\%& \underline{-54.7\%}& \underline{45.7}& -48.7\%& -33.5\%&\underline{+0.7\%}\\
 \gptfour{} &unk.& -\underline{29.7}\%& -46.9\%& -69.2\%& \textbf{-47.4\%}& 36.5 & \textbf{-26.9\%}& -61.8\%&+2.9\%\\
    \bottomrule
    \end{tabular}
    \caption{Overview of the temporal robustness tests. If there is a clear preference, we denote the tests with $\leftrightarrow/\downarrow/\uparrow$ (in table header) to indicate whether we expect well-performing models to be oblivious/decrease/increase in performance on this task. For example, we expect a robust model to be oblivious to relativization and to stay constant in its performance. We report the OE metric for all but event ordering (Contains) and dating (date-match).}
    \label{tab:test_suite_all}
    \vspace*{-2mm}
\end{table*}

\mpara{Evaluation.}
We utilize a set of model-specific metrics (OpenEval and answer equivalence~\cite{kamalloo-etal-2023-evaluating,bulian:2022:emnlp:BEM}) and model-agnostic metrics (i.e., token recall and answer string containment \cite{adlakha:2023:arxiv:evalinstructs,liu:2023:arxiv:longcontextshow,mallen:2023:ACL:whennottrustllms}). 
OpenEval evaluates the correctness of an answer by querying whether a candidate is a suitable answer given the question and the reference answer\footnote{For which we utilize \texttt{Flan-T5-XXL}}. The BEM metric uses a BERT model trained on human-labeled data to predict equivalence between a candidate answer and a reference given a question. For each task, we performed a human alignment study and reported the metric with the highest alignment (Appendix~\ref{sec:appendix_metrics_alignment}) and a full overview of results on all metrics in Appendix~\ref{sec:appendix_al_metrics}.

\section{Testing Temporal Robustness}
\label{sec:results}

In the upcoming sections, we investigate different classes of temporal questions and problems and how our models react to them. For convenience, we show an overview of all models and temporal robustness results in Table~\ref{tab:test_suite_all}.

\subsection{Time Relativization}

The first test that we apply is measuring the effect that switching the time reference from an absolute one (e.g., "2019") to a relative one (e.g., "5 years ago") has on the models' ability to answer temporal factual questions. Relative temporal expressions are a common way to refer to time points, especially when one wants to emphasize the duration of elapsed time. Given that our models can all perform the reasoning needed\footnote{Which we verify by asking, "What year was 5 years ago?"}, one would expect the LLMs to be robust to this paraphrase. Thus, an ideal model should perform on par for both absolute and relative questions (results in Table~\ref{tab:rel_rnd_no_time}).

\begin{table}[ht]
\scriptsize
\centering
\begin{tabular}{l|c|cc|cc} \toprule
\textbf{Model}             &              & \multicolumn{2}{c|}{\textbf{$\leftrightarrow$Relativization}}  & \multicolumn{2}{c}{\textbf{$\downarrow$Removal}}      \\                   & Abs & Abs $\cap$ Rel.    & Diff.& Abs $\cap$ Rem.& Diff.  \\ \hline
 \llamathreeone{} &26.3&13.2&-50.0\%&13.7& -48.0\%\\
 \gemmatwo{} &35.2&25.4&-\textbf{27.9}\%&19.0& -46.1\%\\
 \qwen{} &29.8&17.6&-41.0\%&15.2& -48.9\%\\
 \jamba{} &38.5&\underline{27.3}&-29.1\%&\underline{23.3}& -\textbf{39.5}\%\\
 \commandr{}& \underline{40.4}& 26.6& -34.3\%& \textbf{24.1}& -\underline{40.3}\%\\
 \gptfour{}& \textbf{43.3}& \textbf{30.5}& -\underline{29.7}\%& 23.0&-46.9\%\\
\bottomrule
\end{tabular}
\caption{Results for the relativization and time removal tests measured by the OE metric. We report the intersection between the untransformed (absolute) time references and the two transformations to understand how many of the correct answers are still correct when augmenting the questions to contain, for example, relative time references.}
\label{tab:rel_rnd_no_time}
\vspace{-2mm}
\end{table}

Interestingly, out of the 26.3\% of questions that \llamathreeone{} can answer without paraphrasing, only 13.2\% are also answered correctly when using the relative time reference, resulting in a decrease of 50\%. We observe similar performance decreases in the other models (28-41\%). Specifically, the more capable models seem to be more (but not entirely) robust to using relative references. Given that \textit{all models lack robustness w.r.t. relative time references}, we question how much of the models' performance is due to statistical parroting or a profound and usable understanding of the factual information and the corresponding time component.

\begin{table*}[ht]
\scriptsize
\centering
\begin{tabular}{l|ccc|ccccc|ccc}
\toprule
\textbf{Model} & \multicolumn{3}{c|}{\textbf{$\leftrightarrow$Positioning}}            & \multicolumn{5}{c|}{\textbf{Year Shift (num. of years)}}                    & \multicolumn{3}{c}{\textbf{$\leftrightarrow$Reversal}}           \\
               & Time{[}end{]} & Time{[}front{]} & Diff.         & 0    & 1    & 5    & 10   & Diff.{[}0,10{]} & Fwd    & Fwd $\cap$ Inv & Diff.           \\ \hline
 \llamathreeone{} & 26.3& 27.6& +4.8\%& 26.3& 7.8& 11.0& 7.8& -70.4\%& 7.3& 3.0& -55.6\%\\
 \gemmatwo{} & 35.2& 36.4& +3.3\%& 35.2& 15.2& 20.3& 15.2& -56.9\%& 15.2& 6.0& -60.6\%\\
 \qwen{} & 29.8& 30.0& \textbf{+0.6\%}& 29.8& 9.5& 12.5& 9.6& -68.2\%& 12.1& 3.9& -67.7\%\\
 \jamba{} & 38.5& 39.9& +3.5\%& 38.5& \textbf{21.7}& \textbf{26.0}& \textbf{21.7}& -43.6\%& \underline{23.7}& 9.0& -62.1\%\\
 \commandr{}& \underline{40.4}& \underline{40.7}& \underline{+0.7\%}& \underline{40.4}& \underline{18.5}& \underline{23.4}& \underline{18.5}& -54.2\%& 22.4& \underline{10.2}&\underline{-54.7\%}\\
 \gptfour{}& \textbf{43.1}& \textbf{44.4}& +2.9\%& \textbf{43.1}& 13.3& 17.0& 13.3& -69.2\%& \textbf{31.4}& \textbf{16.5}&\textbf{-47.4\%}\\
 \bottomrule
 \end{tabular}
\caption{Results of the tests for changing the position of the time reference, the effect of shifting the referenced year, and the temporal reversal test. We report OE scores.}
\label{tab:position_error_inverse}
\vspace*{-2mm}
\end{table*}

\subsection{Time Removal}

In the time removal test, we study the relevance of the temporal reference on the question-answering performance. 
To do so, we remove the temporal references from the questions (Table~\ref{tab:rel_rnd_no_time}). 
The model performance decreases by a substantial and surprisingly uniform margin of 40-48\%. 
Conversely, this also means that many temporal questions can be answered (or guessed) correctly without the referenced year's temporal grounding, posing questions about how we evaluate the capture of temporal information in current temporal QA datasets.   

\mpara{What does lower performance mean?}
We think that discarding dates from a question can result in: 
\\
\begin{enumerate*}[label=(\roman*)]
\item the question becoming underspecified and, hence, temporally ambiguous \cite{PiryaniAMJ24}. This means that now answers other than the gold answer $a$ may match the question. In general, several different answers may become correct now besides $a$, as the question can, in principle, refer to any time period. Ideally, LLM should output in this case all the valid answers (or, at least, ask for clarification). In reality, it might just pick one of the answers, likely, the most common one. \\
\item the question becoming more difficult since it is now less informative. This is the case when only one answer $a$ is correct, regardless of the date. A robust LLM should still output the valid answer $a$, or at least ask for more information. 
\end{enumerate*}

Case (i) can arise for questions on \textit{common/repeating types of events} or about \textit{highly dynamic facts}. It is also more likely for shorter questions as they are less specific, resulting in more answers having a match.
Case (ii) may arise for questions related to \textit{specific events} or  \textit{stationary facts}. It is also more likely to happen for more specified questions (where the date is less important as much information is already contained in the question).
Given the dynamic nature of the factual questions discussed in our study, we 
expect decreasing performance after removing temporal references.

\subsection{Time Positioning}
The time positioning test measures the impact of changing the position of the time information within the question on the models' ability to answer time-scoped questions. Specifically, we rewrite the questions, which usually end with the time reference (i.e., "... in 2019?") to instead begin with this time reference (i.e., "In 2019, ..."). To humans, this rewrite of the question should not make a difference, and similarly, we expect models to be robust to these changes (i.e., no change in performance). The results are shown in Table~\ref{tab:position_error_inverse}.

Quite remarkably, \textit{all models benefit from time references to be written at the start of the question}, with relative improvements ranging from 1\% to 5\% in OE score. While it has been intensively studied that language models mostly focus on the first and the last parts of the input while putting less emphasis on the middle part \cite{liu:2023:arxiv:longcontextshow}, this does not fully explain the observations at hand\footnote{We hypothesize reasons in Appendix~\ref{sec:appendix_position}}.

\subsection{Year Shift}

Humans are not always able to remember correct dates. For a question answering, especially over temporal knowledge, to be useful, some lenience regarding errors might be desired. How much exactly is needed and wanted remains to be seen. We include this task on how robust models are to corrupt time references by certain amounts. This task is related to the \textit{Removal} task, but offering slightly wrong dates might still help to anchor the model in the correct time range. We change the years mentioned in the questions to be wrong by $\{0,1,5,10\}$ years.
The results are presented in Table~\ref{tab:position_error_inverse}.
We observe a relatively constant decrease in performance for shifting the years by 1 and 10, but less of a decrease for shifting dates by 5 years. This suggests the existence of repeating events in the date (e.g., such as elections being repeated every 4 years). 

\subsection{Temporal Reversal}

The temporal reversal test is a way to measure the transportability of a fact in another context. Precisely, we test a forward and an inverse formulation of a fact. The forward formulation is the standard question like "Who was the American president in 2005? Answer: Bush". We then reformulate these questions to their inverses, which do not query for the object but for the time of that relation (i.e., "When was Bush the American president? Answer: 2001-2009"). This measures how much the models are susceptible to parroting and how many of these facts are actually understood and usable in differing contexts. The results are shown in Table~\ref{tab:position_error_inverse}.
We notice significant performance drops when looking into questions that were correctly answered in forward \textit{and} inverse forms (47-68\%). This suggests that \textit{many of the correct answers are \textbf{not} due to a sound temporal understanding of the fact}.

\subsection{Temporal Fact Checking}
Next, we evaluate the models' ability to judge the factuality of temporal statements. To do so, we use claims that include temporal statements and measure the degree to which the LLMs can generate the ground-truth labels. This is estimating whether a statement is "True," "False," or if there is "Contradicting" information (Table~\ref{tab:fact_dates_order}).

\begin{table*}[ht]
\scriptsize
\centering
\begin{tabular}{l|cc|cccc|ccccccc} \toprule
\textbf{Model} & \multicolumn{2}{c|}{\textbf{$\uparrow$Fact Checking}}& \multicolumn{4}{c|}{\textbf{$\leftrightarrow$Event Dating}}                                         & \multicolumn{7}{c}{\textbf{$\leftrightarrow$Event Ordering}}                                                                  \\
               & Cont.&OE& Day & Month & Year & Diff.{[}Y,D{]} & 0 & 1 & 5 & 10 & 30 & 100 & Diff.{[}100,0{]} \\ \hline
 \llamathreeone{} & 29.1& 34.2& 13.0& 17.9& 52.8& -75.4\%& 49.9& 51.6& 50.8& 51.6& 51.0& 49.8&\textbf{-0.2\%}\\
 \gemmatwo{} & 39.9& 42.6& 32.9& \underline{49.9}& 70.5& \underline{-42.6\%}& 36.3& 49.4& 36.8& 49.4& 42.1& 49.4& -36.1\%\\
 \qwen{} & \textbf{74.7}& 31.4& 24.6& 22.2& 62.9& -60.9\%& 48.4& \textbf{81.1}& \underline{57.5}& \textbf{81.1}& \underline{74.4}& \underline{81.1}& -67.6\%\\
 \jamba{} & \underline{65.5}& \textbf{52.5}& 36.0& 30.9& 75.4& -52.3\%& 49.0& 49.5& 50.2& 49.5& 51.5& 49.5& \underline{-1.0\%}\\
 \commandr{}& 46.5&\underline{45.7}& \underline{38.7}& 47.2& \underline{75.5}& -48.7\%& \underline{52.8}& 50.6& 54.4& 57.8& 64.1& 70.4&-33.5\%\\
 \gptfour{}& 33.1&36.5 & \textbf{56.3}& \textbf{56.3}& \textbf{77.0}& \textbf{-26.9\%}& \textbf{53.5}& \underline{56.9}& \textbf{64.0}& \underline{70.1}& \textbf{79.4}& \textbf{86.6}&-61.8\%\\     
\bottomrule
\end{tabular}
\caption{Results for the fact checking, event dating, and ordering tasks. We report the date match metric (event dating) and Contains (event ordering). 0,1,5,10,30,100 is the distance in years between the compared events.}
\label{tab:fact_dates_order}
\vspace*{-2mm}
\end{table*}

Given that this task is a three-class classification problem, the results of all models are lacking. Interestingly, we observe lower performance for the most capable \gptfour{}. Upon manual analysis, many models avoid answering questions for their lack of information. This might result from their training, resulting in better-calibrated models and lower performance on this task.

\begin{table*}[ht]
    \scriptsize
    \centering
    \begin{tabular}{l|l|l|c} \toprule
    \textbf{Test}     & \textbf{Question}                                             & \textbf{Prediction}            & \textbf{Consistent /w orig. Prediction} \\ \hline
    Original & Bernardo Corradi played for which team in 2006?      & Fiorentina            &                                \\ \hline
    Relativization     & Bernardo Corradi played for which team 17 years ago? & Inter Milan           & \xmark                              \\
    Removal  & Bernardo Corradi played for which team?              & Italian National Team & \xmark                              \\
    Positioning & In 2006, Bernardo Corradi played for which team?     & No answer             & \xmark                              \\
    Reversal  & When did Bernardo Corradi play for Fiorentina?       & He never did          & \xmark                              \\ \bottomrule        
    \end{tabular}
    \caption{Example of the automatic test suite. For a given question, we automatically create paraphrases inspired by the temporal robustness tests and retrieve answers from the model. Looking into the consistency between the answer to the original question and the test questions can help us judge how much model predictions can be trusted.}
    \label{tab:auto_test_suite}
    \vspace*{-2mm}
\end{table*}

\subsection{Event Dating}

We next use events from Wikipedia year pages\footnote{E.g., \url{https://en.wikipedia.org/wiki/2009}} and predict their date in day, month, and year precision. In our prompt, we specify a format in which we would like to receive the dates ("dd-mm-yyyy"), but we observe many models not adhering to the format. While this is not critical as long as the answer is correct, measuring correctness with the metrics at hand becomes difficult. The contains metric fails understandably when "11-11-1995" becomes "11th of November 1995" and we also observe that BEM and OpenEval not to be robust in terms of how dates are phrased. 
Based on different formats used by the models, we built our own date-matching metric\footnote{details in Appendix~\ref{sec:appendix_datematch}} and report the event dating performance in Table~\ref{tab:fact_dates_order} (middle).

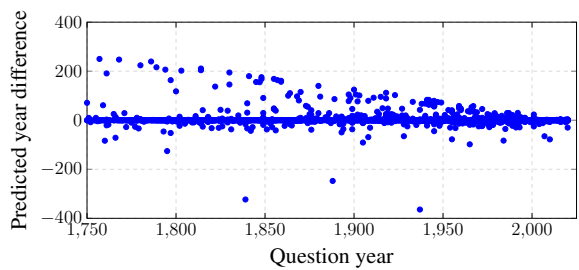
\begin{figure}[ht]
    \begin{center}
            \scalebox{.48}{\begin{tikzpicture}
    \begin{axis}[
        width=15cm,
        height=7cm,
        xlabel={\LARGE Question year},
        xmin=1750, xmax=2025,
        xtick style={color=black},
        xtick={1750,1800,1850,1900,1950,2000},
        ylabel={\LARGE Predicted year difference},
        ymin=-400, ymax=400,
        ytick={-400, -200, 0, 200, 400},
        grid=both,
        grid style={dashed, gray!30},
        tick label style={font=\Large},
    ]
        \addplot[
            only marks,
            mark=*,
            mark size=2pt,
            color=blue,
        ] table [col sep=comma] {Figures/event_dating_years_data.csv};
    \end{axis}
\end{tikzpicture}}
    \end{center}
    \setlength{\abovecaptionskip}{3pt}
    \caption{Difference between ground-truth and predicted years for \llamathreeone{}.}
    \label{fig:event_dating_years}
    \vspace*{-2mm}
\end{figure}

As expected, we find the larger models to outperform the smaller models. 
Also, we observe that \textit{years, as the coarsest granularity, are better captured than months or days}. Going from year-to-day precision, performance drops by 27\% or more, suggesting that LLMs "know" temporal facts up to a certain precision. 
The date-matching metric also allows us to measure the difference between predicted and ground-truth dates. We plot the deviation between \llamathreeone{}'s predicted and ground-truth years in Figure~\ref{fig:event_dating_years}.

While we observe that most predictions are somewhat close to the actual year (53\% even matching it precisely), we also find many questions answered with rather distant years. 
The worst predicted year was wrong \textit{by over 300 years} (c.f. Figure~\ref{fig:event_dating_years}). 
Interestingly, \llamathreeone{} tends to output too recent dates for many questions.

\subsection{Event Ordering}

Our last test is whether LLMs can order events chronologically. To do so, we again use the major events from the Wikipedia year pages and always pass two events with the question asking whether A happened before B. This will shed additional light on whether the models have a linear understanding of time and whether this is present, usable information to them. While it has been shown that \texttt{Chat-GPT} performs well at event detection and reasoning about causal relationships, it seems not to be proficient at identifying temporal order in the case or discourse analysis \cite{chan:2023:arxiv:chatgpttemporalcausaldisscourse}. 
The results of the event ordering are in Table~\ref{tab:fact_dates_order} (right). 

First and foremost, we find \llamathreeone{}, \gemmatwo{}, and \jamba{} to have consistently bad performance similar to majority classifiers ($\sim50\%$). For the remaining models, we see the performance increase the further apart the events. 
Typically, one would expect the models, given they are quite capable of dating events w.r.t. years, but less so w.r.t. days, to perform worse at ordering events from the same year than events that happened 1, 5, or 10 years apart. 
Even for events from 30 or 100 years apart, where humans are likely to infer the correct order, we only observe a maximum improvement of 26\% (\gptfour{}). 
Unlike humans, who find it natural to differentiate between events that are long time gaps apart, LLMs find it hard to order events despite successfully dating them.

\section{Automatic Robustnesss Testing}
\label{sec:automatic_tests}

As with all test suites, our tests for temporal robustness, as discussed in the previous sections, are built a priori and require that we \textit{know the answers} to the questions. However, many of our tests are query-centric reformulations that can be applied to almost generic questions (ending in a temporal reference like ``in 2007?''). Thus, we wonder whether we could use these query-centric tests to better judge if the model correctly processed and understood the time component of the question. In this setting, we are \textit{not required} to know the ground-truth answer, which allows us to use this automatic test suite on the fly. This can be beneficial either when hosting the chat/question-answering system to understand when the models might output wrong answers to users or by directly showing the results so that users may use this additional information as contextualization to decide whether they want to trust the model answer or not (example in Table~\ref{tab:auto_test_suite}).

Given the lack of robustness in the relativization, positioning, and reversal tests, one might question any model's temporal literacy. 
Note that the model might still produce the correct answer, even if the predictions and their consistency might lead us to believe it is not sure about this. 
However, if the model produces the correct answer, it is more likely due to shortcuts or chance and not because it understood the temporal factual relation.

Next, we show that the query-centric reformulations can be used to estimate whether LLM answers are correct.
We use $3,000$ questions for which we have ground-truth answers, creating automatic question reformulations and getting the model predictions for all questions. We evaluate the consistency between the original and test question predictions, resulting in four values of 0 to 1. We find these four consistency scores to be predictive of the actual correctness of the answer, outperforming a majority classifier on a balanced test set by $14.9\%$ (\llamathreeone). This result emphasizes that besides better understanding and calibrating user trust in the predictions, these \textit{scores can help to evaluate the correctness of model predictions automatically}.

\section{Reformulations for Better QA}
\label{sec:findings_application}

Lastly, we take another approach to validate the findings from the robustness test. We directly apply the findings about lacking robustness to questions to improve the LLMs for temporal QA. While this was not the main intention of our study, applying the findings emphasizes the usefulness of the study's results. We sample 1k previously unused questions from ArchivalQA and apply the following transformations. First, we remove the time to simulate the effectiveness of adding time references in case they are missing. In subsequent steps, we move from relative to absolute time references and move the temporal reference to the front of the question. The results are shown in Table~\ref{tab:findings_test}.

\begin{table}[h]
\centering
\scriptsize
\begin{tabular}{lcccc} \toprule
\textbf{Model}                       & \commandr{} & \gemmatwo{} & \gptfour{}         &Avg. Gain\\ \midrule
$Q_{No\,\,Time}$                              &                      30.0&                    20.6&                    27.8& --\\
$Q_{+\,\,Relative}$      &                      31.9&                    27.4&                    32.3&+18.5\%\\
$Q_{+\,\,Absolute}$     &                      39.2&                    35.7&                    42.2&+28.0\%\\
$Q_{+\,\,Time[front]}$ & 40.9
&                    35.5&    43.5&+2.3\%\\  \bottomrule             
\end{tabular}
\caption{We apply the findings of our study into LLM temporal robustness to new questions. By applying automatic transformations to questions in forms to which the LLMs are not robust to (e.g., relative to absolute references), we can improve the QA performance. Values in OE scores.}
\label{tab:findings_test}
\vspace*{-2mm}
\end{table}

Moving from no time reference to relative time offers an average performance boost of 18.5\%. Similarly, rewriting relative to absolute references and moving absolute references to the front offer an improvement of 28.0\% and 2.3\%, respectively. The total improvement of adding absolute time references to the front of questions, which had no time reference, would be 55\% on average.

\section{Discussion \& Conclusion}
\label{sec:conclusion}

The temporal robustness tests presented in this work offer a first suite to benchmark temporal processing abilities of LLMs. Besides actionable insights -- like avoiding relative time references for most models or starting temporal questions with a time reference -- we offer tests that help investigate what temporal understanding is present in models.   

The temporal robustness tests may be used in addition to the typical task performance as pre-deployment checks to evaluate models' abilities and problems better. For example, one might take \gptfour{} and \jamba{}, which perform very similarly on temporal QA (absolute values in Table~\ref{tab:rel_rnd_no_time}) and take a closer look at our detailed tests to understand that \jamba{} is less robust to temporal reversal (-15\%) and worse at dating events in day-precision (-25\%), but more proficient at judging temporal factual claims (+16\%). 

In this study, we examined the temporal robustness of LLMs. We tested a variety of LLMs using a suite of eight tests assessing different kinds of temporal abilities and robustness to natural paraphrases of questions. While we generally observe higher QA performance for bigger models such as \gptfour{}, we did not find these models robust to our temporal robustness tests. This study serves as the inaugural benchmark for LLMs' temporal robustness, providing valuable insights into correct temporal information processing and model failures.  
Further, we found our temporal robustness tests applicable along the entire model lifecycle: 1) For developers to benchmark their models and understand which abilities need improvement. 2) As pre-deployment checks to understand the differences between similarly performing models. 3) By using our automatic tests to help users gauge whether or not to trust the model's predictions. 4) By guiding question reformulations for improved QA performance. We believe this set of tests to help study the temporal robustness of LLMs.

\section*{Limitations}
\paragraph{Are these tests a comprehensive set?}
 We deal with real-world events, and absolute dates matter; hence, we use several date interventions/perturbations. 
Temporal ordering and dating are well understood in the literature on events in the Web and IR community~\cite{tran2015timeline,campos2021automatic:timeline,setty2017modeling,bradburn1999temporal:ordering}.
We leave out operations that belong to temporal algebra \cite{karger:1998:math:temporal_algebra} or temporal logic \cite{konur:2013:frontiers:survey_temporal_logic}
because we are not interested in arbitrary temporal operations and rather focus on events.
We are also focused on LLM-based mistakes; our reformulations are natural and not adversarial. Therefore, each transformation is a plausible question, and we leave out all adversarial reformulations.
Having made these assumptions (based on the useful and real-world character of the QA task), we acknowledge this is the first step that could -- in the future -- be extended by rich temporal scopes such as date spans and before \& after relations.

\paragraph{What about retrieval?}
Using a retrieval system and adding additional information is a logical step when striding toward effective QA or chat systems. However, we focus on the innate abilities and temporal robustness of LLMs. When we add context containing the answer to a question, the problem changes from recalling factual information and handling time to a reading comprehension problem -- and LLMs are quite proficient at these. The critical part in the retriever setting is retrieving the correct information, which is by no means trivial, especially when discussing historical information that might be relatively rare or incomplete.
We deem the retrieval setting to be out of this work's scope.   

\paragraph{Do QA metrics work for temporal questions?}
As discussed in Section~\ref{sec:study} and Appendix~\ref{sec:appendix_metrics_alignment} not all metrics work equally well for all temporal tasks in this study. Specifically, handling dates and time occurrences with their various surface forms was not robust by the existing metrics. Therefore, exploring temporal QA and IR metrics might be a worthwhile endeavor for future work.

\section*{Ethics Statement}
We observe the brittle behavior of LLMs for temporal factual questions. While this may be used to achieve sub-optimal performance, we do not believe this is a directly suitable attack vector to achieve harmful behavior. While not directly derivable from this work, it might be possible to use adversarial attacks to intentionally bias the outputs of LLMs for temporal questions, given the brittle behavior showcased in this study. If successful, this could result in LLMs outputting fake historical information.

\section*{Acknowledgments}
This work was supported by the Lower Saxony Ministry of Science and Culture (MWK), in the zukunft.niedersachsen program of the Volkswagen Foundation (HybrInt).

\bibliographystyle{Template/acl_natbib}
\bibliography{references}

\appendix
\section{General Questions}
\subsection{Did you describe the limitations of your work?}
yes, see Limitations

\subsection{Did you discuss any potential risks of your work?}
Yes, see Ethics Statement

\section{Scientific Artifacts}

\subsection{Did you cite the creators of artifacts you used?}
Yes, see Section~\ref{sec:study}

\subsection{Did you discuss the license or terms for use and / or distribution of any artifacts?}

\begin{enumerate}
    \item ArchivalQA: Apache 2.0, \url{https://github.com/WangJiexin/ArchivalQA}
    \item Temporal Facts (Quantemp): CC BY-NC 4.0, \url{https://github.com/factiverse/QuanTemp}
    \item Time-Sensitive QA: BSD 3-Clause, \url{https://github.com/wenhuchen/Time-Sensitive-QA}
    \item Wikipedia Year pages: CC BY-SA 4.0, e.g., \url{https://en.wikipedia.org/wiki/2006} 
\end{enumerate}

We do not plan to distribute these artifacts ourselves but provide scripts to construct the data used in the paper.

\subsection{Did you discuss if your use of existing artifact(s) was consistent with their intended use, provided that it was specified? For the artifacts you create, do you specify intended use and whether that is compatible with the original access conditions (in particular, derivatives of data accessed for research purposes should not be used outside of research contexts)?}

The used artifacts specify non-commercial use. Our usage was consistent with their specifications.

\subsection{Did you discuss the steps taken to check whether the data that was collected / used contains any information that names or uniquely identifies individual people or offensive content, and the steps taken to protect / anonymize it?}

We only collect data from Wikipedia year pages. This may contain names of public figures such as presidents, government figures, or prominent people, and therefore, did not anonymize their names.

\subsection{Did you provide documentation of the artifacts, e.g., coverage of domains, languages, and linguistic phenomena, demographic groups represented, etc.?}

We cover an overview of the used artifacts in Section~\ref{sec:study} as well as Appendix~\ref{sec:appendix_source_ds}. 

\subsection{ Did you report relevant statistics like the number of examples, details of train / test / dev splits, etc. for the data that you used / created?}

We cover an overview of the used artifacts in Section~\ref{sec:study} as well as Appendix~\ref{sec:appendix_source_ds}.

\section{Computational Experiments}

\subsection{Did you report the number of parameters in the models used, the total computational budget (e.g., GPU hours), and computing infrastructure used?}

The model parameters are listed in Table~\ref{tab:models}. We did not train or fine-tune models, but ran inference on a larger set of models. Our used infrastructure was a mixture of A100 with 40/80GB memory. Running the entire test suite may take ca. 1 day on one GPU per model, resulting in 6 GPU/days for all models combined.

\subsection{Did you discuss the experimental setup, including hyperparameter search and best-found hyperparameter values?}

We did not run a hyperparameter search, but we described our experimental setup both in Section~\ref{sec:study} and made our code available at \url{https://anonymous.4open.science/r/temporalrobustness-B3D3/}. The repository contains hyperparameters, prompts, etc.

\subsection{Did you report descriptive statistics about your results (e.g., error bars around results, summary statistics from sets of experiments), and is it transparent whether you are reporting the max, mean, etc. or just a single run?}

We are reporting single-run results since this study contains descriptive results and does not try to show a clear benefit of using one model over another. 

\subsection{If you used existing packages (e.g., for preprocessing, for normalization, or for evaluation, such as NLTK, Spacy, ROUGE, etc.), did you report the implementation, model, and parameter settings used?}

Details available at: \url{https://anonymous.4open.science/r/temporalrobustness-B3D3/}

\section{Human Annotators}

\subsection{Did you report the full text of instructions given to participants, including, e.g., screenshots, disclaimers of any risks to participants or annotators, etc.?}

N/A, did not use human annotators

\subsection{Did you report information about how you recruited (e.g., crowdsourcing platform, students) and paid participants, and discuss if such payment is adequate given the participants’ demographic (e.g., country of residence)?}

N/A, did not use human annotators

\subsection{Did you discuss whether and how consent was obtained from people whose data you’re using/curating?}

N/A, did not use human annotators

\subsection{Was the data collection protocol approved (or determined exempt) by an ethics review board?}

N/A, did not use human annotators

\subsection{Did you report the basic demographic and geographic characteristics of the annotator population that is the source of the data?}    

N/A, did not use human annotators

\section{Use of AI Assistants}

\subsection{Did you include information about your use of AI assistants?}

We did not use AI assistants to generate text or perform research directly. We did use ChatGPT and Grammarly to perform reformulations of existing text as well as to fix grammatical errors.

\label{section:appendix}

\section{Additional Setup Details}
\label{sec:appendix_implementation_details}

\subsection{Extended Discussion of Temporal Source Datasets}
\label{sec:appendix_source_ds}
\mpara{Time-sensitive-QA dataset}~\cite{chen:2021:NeurIPS:timesensitiveqa} is constructed by mining time-evolving facts from WikiData and aligning them to their corresponding Wikipedia pages, employing crowd workers to verify and calibrate noisy facts, and generating question-answer pairs based on the annotated time-sensitive facts. 
The dataset contains 40,000 question-answer pairs focusing on around 5,500 time-evolving facts; it is structured into two variants based on difficulty: easy and hard.

\mpara{TemporalQuestions dataset}~\cite{wang2021improving} is designed to evaluate the capability of QA systems to handle time-scoped questions. This dataset focuses on questions related to specific events and their temporal aspects, derived from historical news archives and other temporally rich sources. The dataset contains 1,000 human-generated questions about major events, half of which are explicitly and half implicitly time-scoped, meaning half of the questions contain temporal expressions. In contrast, the remaining ones lack any temporal references.
 
\mpara{ArchivalQA}~\cite{DBLP:conf/sigir/WangJY22} is a large-scale collection designed specifically for temporal news QA, containing 532,444 question-answer pairs, often on detailed and minor aspects. These pairs are derived from the New York Times Annotated Corpus, which spans news articles published between January 1, 1987, and June 19, 2007. The dataset-constructing framework with automatic question generation and filtering steps ensures high-quality and non-ambiguous questions.

\subsection{Additional Model Details}
\label{sec:appendix_model_details}
\begin{table*}
    \centering
    \scriptsize
    \begin{tabular}{l|l|l|l}
        \toprule
        \textbf{Model Name} & \textbf{Mode Size} & \textbf{Notes} & \textbf{ Cutoff} \\
        \hline

 \llamathreeone{}& 8B& Instruction-tuned version&Dec. 2023\\
 \jamba{}& 12B active, 52B total& Mixture-of-Experts model that combines mamba (state-space) and transformer blocks. 8bit quant.&Mar. 2024\\
 \gemmatwo{}& 27B& Instruction-tuned version&Jun. 2024\\
 \qwen{}& 32B& Instruction-tuned version&2023\\

        \commandr{}& 104B params & RAG-optimized language model, weights openly available. Uses 4-bit quantization & N.S.\\
        
        \gptfour{}& N.S. & OpenAI's flagship GPT model (gpt-4-1106-preview)& Apr. 2023\\
        \bottomrule
    \end{tabular}
    \caption{Summary of different models with their respective details}
    \label{tab:models}

\end{table*}

As shown in Table~\ref{tab:models}, we use a selection of competitive LLMs. Specifically, we use thee following versions: \llamathreeone{}\footnote{\url{https://huggingface.co/meta-llama/Llama-3.1-8B-Instruct}}, \gemmatwo{}\footnote{\url{https://huggingface.co/google/gemma-2-27b-it}}, \jamba{}\footnote{\url{https://huggingface.co/ai21labs/AI21-Jamba-1.5-Mini}}, \qwen{}\footnote{\url{https://huggingface.co/Qwen/Qwen2.5-32B-Instruct}}, \commandr{}\footnote{\url{https://huggingface.co/CohereForAI/c4ai-command-r-plus-4bit}}, and \gptfour{}\footnote{\url{https://platform.openai.com/docs/models/gp\#gpt-4-turbo-and-gpt-4}}.

\subsubsection{Temporal QA Performance}
We evaluate the temporal robustness of LLMs in the study. Still, for completeness reasons, we also provide the downstream QA performance of our models on three established temporal QA benchmark datasets. The results are shown in Table~\ref{tab:temporalQA}. The performance of all models, on all metrics, leaves an opportunity for improvement.

\begin{table*}[ht]
    \scriptsize
    \centering
    \begin{tabular}{lcccc|cccc|cccc}
    \toprule
    \textbf{Model}              & \multicolumn{4}{c|}{\textbf{ArchivalQA}}                                                                                          & \multicolumn{4}{c|}{\textbf{TemporalQuestions}}                                                                                            & \multicolumn{4}{c}{\textbf{Time-Sensitive-QA}}                                                                                             \\
                                & Recall                         & Cont.                          & BEM                            & OE                             & Recall                         & Cont.                          & BEM                            & OE                             & Recall                         & Cont.                          & BEM                            & OE                             \\ \midrule
 \llamathreeone{} & 22.2& 18.5& 41.4& 23.2& 63.1& 58.9& 76.8& 62.1& 13.3& 7.7& 30.7&13.7\\
 \jamba{} & 30.0& 24.3& \textbf{49.9}& 35.4& 73.3& 68.6& \textbf{86.9}& 74.2& 28.6& 18.2& 44.7&32.3\\
 \gemmatwo{} & 29.3& 25.0& 40.6& 32.3& 76.8& 72.2& 85.7& 78.4& 23.4& 16.0& 32.4&26.0\\
 \qwen{} & 25.6& 21.1& 31.6& 26.0& 63.6& 60.1& 72.1& 64.4& 19.5& 10.0& 18.6&16.5\\
    \commandr{}                  & 31.8                           & 26.1                           & 46.5                   & 37.0                           & 76.3                           & 72.1                           & 81.5                           & 79.1                           & 30.7                           & 21.1                           & 38.6                           & 32.7                           \\
    \gptfour{}               &                                \textbf{38.5}&                                \textbf{32.8}&                                46.3&                                \textbf{39.8}&                                \textbf{81.7}&                                \textbf{76.3}&                                86.7&                                \textbf{81.9}&                                \textbf{44.2}&                                \textbf{33.2}&                                \textbf{46.3}&                                \textbf{39.3}\\ \bottomrule
    \end{tabular}
    \caption{Performance of our models on common temporal factual QA benchmarks.}
    \label{tab:temporalQA}
\end{table*}

\subsubsection{Effect of Prompts}
We note that most of our study uses the standard prompts and did not include any prompt engineering or established best practice (e.g., Chain-of-Thought \cite{wei:2022:neurips:cot} or role-playing\footnote{"You are a historian [...]"} \cite{kong:2024:arxiv:roleplayingprompts}) prompts. We also experiment with these two best practice prompts\footnote{All prompts will be made available with our code.} and show the results for a selection of models in Table~\ref{tab:qa_prompts}. While the historian role-playing prompt performs competitively across the board, the CoT prompt does not and might be unsuitable for factual recall, which usually might not involve multi-step reasoning. Lastly, we expect prompt tuning to improve the overall model performance. Still, we did not see clear evidence that better-performing models consistently outperform others in robustness to our temporal paraphrases.

\begin{table*}[ht]
    \small
    \begin{tabular}{l|c|cccc|cccc|cccc}
    \toprule
    \textbf{Model}                 & \textbf{Prompt} & \multicolumn{4}{c}{\textbf{ArchivalQA}} & \multicolumn{4}{c}{\textbf{TemporalQuestions}} & \multicolumn{4}{c}{\textbf{Time-Sensitive hard QA}} \\
                                   &                 & Recall    & Cont.    & BEM    & OE   & Recall      & Cont.      & BEM     & OE     & Recall      & Cont.      & BEM     & OE     \\ \hline
    \multirow{3}{*}{\commandr{}}& default &  31.8                           & \textbf{26.1}                           & 46.5                   & \textbf{37.0}                          & \textbf{76.3}                           & \textbf{72.1}                           & 81.5                           & \textbf{79.1}                           & 30.7                           & 21.1                           & 38.6                           & 32.7                           \\ 
                                   & CoT  &  12.9&  12.2& 20.5& 35.2& 34.3&  30.9& 43.7& 71.0& 13.6& 8.3& 16.8& \textbf{43.8}\\
                                   & historian       &  \textbf{32.5}&  \textbf{26.1}& 46.4& 36.0&  74.7&   69.7&    \textbf{82.2}&  77.2&    \textbf{32.5}&  \textbf{21.9}&   \textbf{38.8}&  33.6\\ \hline
 \multirow{3}{*}{\jamba{}}& default& 30.0& 24.3& \textbf{49.9}& 35.4& 73.3& 68.6& 86.9& 74.2& 28.6& 18.2& 44.7&32.3\\
 & CoT& 29.9& 24.4& 49.7& 35.1& 73.2& 68.5& \textbf{87.6}& 72.5& 28.5& 18.3& 45.5&32.9\\
 & historian& \textbf{30.6}&\textbf{25.0}& 47.5& \textbf{36.1}& \textbf{74.7}& \textbf{69.6}& 83.9& \textbf{76.7}& \textbf{30.3}& \textbf{20.0}& \textbf{46.4}&\textbf{34.5}\\ \hline
 \multirow{3}{*}{\llamathreeone{}}& default& 22.2& 18.5& \textbf{41.4}& 23.2& 63.1& 58.9& \textbf{76.8}& 62.1& 13.3& 7.7& \textbf{30.7}&13.7\\
 & CoT& 5.2& 3.3& 10.3& \textbf{36.1}& 11.6& 8.9& 18.8& \textbf{71.8}& 7.9& 3.3& 11.9&\textbf{33.8}\\
 & historian& \textbf{23.0}& \textbf{19.2}& 41.1& 23.6& \textbf{64.4}& \textbf{60.1}& 76.2& 60.9& \textbf{15.5}& \textbf{9.8}& 29.9&15.1\\
   \bottomrule
    \end{tabular}
    \caption{Overview of the temporal QA of our models when using different prompting schemes.}
    \label{tab:qa_prompts}
\end{table*}

\section{More discussion on the Effect of Positioning Time}
\label{sec:appendix_position}
Given that we employ a system prompt and a prompt that specifically asks for the following question to be answered, the time reference at the front of the question is hardly at the beginning of the model input. However, the time reference at the end is almost at the end of the input and should, therefore, be focussed on by the model. Yet, the performance is found to be superior when the time reference is before the question. 
We hypothesize that a different thing is at play here: The residual stream does not have enough bandwidth to store all historical information on certain entities and relations. \citet{meng:2022:neurips:locatingandeditingfactual} found that when answering factual questions about entities, the embeddings of the last entity token would be enriched with as much information as possible on that entity by retrieving it from the feed-forward layers. This information is then copied to the last token embedding, where the attention mechanism selects the information necessary to answer the question from the embedding.

Let us look at an example in our normal question form: 
\emph{Who was the American president in  2019?} %
Remember that all LLMs in this study are autoregressive language models (i.e., their attention may only look at the previous context of a given token). We see that the token "president," in which the factual information will be aggregated, has no "understanding" that it needs to find information on what was the case in 2019. Therefore, it would either save the most recent information or try to aggregate all information. Using the most recent information will likely fail with historical knowledge (and given our other results, we do not believe this to be the case). Trying to enrich the embedding with all information on the American president might fail because there is too much information. When the last token's attention then tries to retrieve the correct information about our year, it might not be accessible, and the question might be answered incorrectly. However, if the question starts with the time--reference, the entity token ("president") can be precisely enriched with the information from the correct years and information from other years may be then de-prioritized or discarded. While this hypothesis needs to be thoroughly tested, for our case of measuring temporal robustness, we can conclude that the desired output would be models that are robust to where the time references occur in the question. When aiming for the best QA performance, however, this result suggests formulating temporal questions to start with their temporal references.%

\section{Do Metrics align with Human Judgement?}
\label{sec:appendix_metrics_alignment}
To check whether our metrics align with human judgment, we randomly sampled 100 QA pairs from \gemmatwo{} and manually scored every answer. Table~\ref{tab:metrics_alignment} presents the resulting correlation and agreement scores. Generally, all metrics align well with human judgment with agreement rates of 87\% and above.

\begin{table*}[ht]
    \centering
    \small
    \begin{tabular}{ll|c|c|c|c}
        \toprule
         \textbf{Task}&\textbf{Metric}& \textbf{Matthews Correlation} & \textbf{Pearson Correlation} & \textbf{Cohen Kappa} & \textbf{Agreement Rate} \\
        \hline
         &\textbf{Cont.} & 0.7522 & 0.7522 & 0.7420 & 0.89 \\
         \textbf{QA}&\textbf{BEM} & 0.7322 & 0.7322 & 0.7241 & 0.87 \\
         &\textbf{OE} & \textbf{0.9778} & \textbf{0.9778} & \textbf{0.9776} & \textbf{0.99} \\
        \hline
 & \textbf{Cont.} & 0.757& 0.757& 0.729&0.88\\
 \textbf{Fact Checking}& \textbf{BEM}& 0.391& 0.391& 0.307&0.62\\
 & \textbf{OE}& \textbf{0.826}& \textbf{0.826}& \textbf{0.811}&\textbf{0.92}\\ \hline
 & \textbf{Cont.} & 1.0& 0.9999& 1.0&1.0\\
 \textbf{Event Dating}& \textbf{BEM}& 0.6624& 0.6624& 0.6099&0.8\\
 & \textbf{OE}& 1.0& 0.9999& 1.0&1.0\\ 
 & \textbf{date-match}& 1.0& 0.9999& 1.0&1.0\\ \hline
 & \textbf{Cont.}& \textbf{1.0}&\textbf{ 1.0}& \textbf{1.0}&\textbf{1.0}
\\
 \textbf{Event Ordering}& \textbf{BEM}& 0.081& 0.081& 0.080&0.54\\
 & \textbf{OE}& 0.0& n.d& 0.0&0.5\\ \bottomrule
    \end{tabular}
    \caption{Correlation of QA metrics with human judgment on 100 (QA) and 50 (remaining tasks)  randomly sampled data points.}
    \label{tab:metrics_alignment}
\end{table*}

\section{Results for other Metrics}
\label{sec:appendix_al_metrics}

The following tables contain results on other metrics not presented in the main content of the paper for space limitations. 

Table~\ref{tab:rel_rnd_no_time_all} contains the results for the relativization and removal tests.  
\begin{table*}[ht]
\scriptsize
\centering
\begin{tabular}{l|ccc|ccc|ccc|ccc|ccc}
\toprule
\textbf{Model}                & \multicolumn{3}{|c|}{}                           & \multicolumn{6}{c|}{\textbf{$\leftrightarrow$Relativization}}                                                     & \multicolumn{6}{c}{\textbf{$\downarrow$Removal}}                                    \\ \hline
                              & \multicolumn{3}{c|}{\textbf{Abs}}                        & \multicolumn{3}{c|}{\textbf{Abs $\cap$ Rel.}}        & \multicolumn{3}{c|}{\textbf{Diff.}}                                  & \multicolumn{3}{c|}{\textbf{Abs $\cap$ Rem.}}        & \multicolumn{3}{c}{\textbf{Diff.}}     \\ \hline
                              & Cont.     & BEM                           & OE & Cont. & BEM                           & OE & Cont. & BEM                                           & OE & Cont. & BEM                           & OE & Cont. & BEM              & OE \\ \hline
\llamathreeone{} &           21.2& 39.2                          &    26.3&       10.7& 22.3                          &    13.2&       -49.6\%& -43\%                              &    -50.0\%&       10.3& 19.1                          &    13.7&       \textbf{-51.5\%}& \textbf{-51\%}&    \underline{-48.0\%}\\
\gemmatwo{}      &           27.4& 40.6                          &    35.2&       \underline{19.8}& 27.8                          &    25.4&       \textbf{-27.5\%}& -\underline{32}\% &    \textbf{-27.9\%}&       14.5& 20.6                          &    19.0&       -46.9\%& -49\% &    -46.1\%\\
\qwen{}          &           24.8& 32.9                          &    29.8&       14.5& 19.2                          &    17.6&       -41.7\%& -42\%                              &    -41.0\%&       12.8& 16.4                          &    15.2&       \underline{-48.3\%}& \underline{-50\%}&    \textbf{-48.9\%}\\
\jamba{}         &           27.0& \underline{47.0} &    38.5&       18.7& \textbf{37.1}                 &    \underline{27.3}&       \underline{-30.9\%}& -\textbf{21}\%    &    \underline{-29.1\%}&       15.5& \textbf{29.2}                 &    \underline{23.3}&       -42.8\%& -38\% &    -39.5\%\\
\commandr{}      &  \underline{28.6}& \textbf{47.5}                 &    \underline{40.4}&       11.0& 30.9                          &    26.6&       -61.5\%& -35\%                              &    -34.3\%&       \underline{16.5}& \underline{26.7} &    \textbf{24.1}&       -42.4\%& -44\% &    -40.3\%\\
\gptfour{}       &  \textbf{35.6}& \textbf{47.5}                 &    \textbf{43.3}&       \textbf{23.9}& \underline{32.2} &    \textbf{30.5}&       -33.3\%& -\underline{32}\% &    -29.7\%&       \textbf{19.3}& 25.9                          &    23.0&       -45.9\%& -46\% & -46.9\%\\  
\bottomrule
\end{tabular}
\caption{Results for the relativization and time removal tests measured by multiple metrics (Contains, BEM, and OE).}
\label{tab:rel_rnd_no_time_all}
\end{table*}

Table~\ref{tab:position_error_all} contains the results for the positioning and year shift tests. 
\begin{sidewaystable}[ht]
\scriptsize
\centering
\begin{tabular}{l|ccc|ccc|ccc|ccc|ccc|ccc|ccc|ccc|}
\toprule
\textbf{Model} &  \multicolumn{9}{c|}{\textbf{$\leftrightarrow$Positioning}}&  \multicolumn{15}{c}{\textbf{Year Shift (num. of years)}}\\ \hline
               &  \multicolumn{3}{c|}{Time{[}end{]}}&  \multicolumn{3}{c|}{Time{[}front{]}}&  \multicolumn{3}{c|}{Diff.}&  \multicolumn{3}{c|}{0}&  \multicolumn{3}{c|}{1}&  \multicolumn{3}{c|}{5}&  \multicolumn{3}{c|}{10}&  \multicolumn{3}{c|}{Diff.{[}0,10{]}}\\ \hline
 &  Cont. & BEM&OE&  Cont.& BEM&OE&  Cont.& BEM&OE&  Cont.& BEM&OE&  Cont.& BEM&OE&  Cont.& BEM&OE&  Cont.& BEM&OE&  Cont.& BEM&OE\\ \hline

 \llamathreeone{} &  21.2&39.2 &26.3&  22.0&52.5 &27.6&  +3.8\%&+34\% &+4.8\%&  21.2&39.2 &26.3&  12.3&29.0 &7.8&  15.4&32.9 &11.0&  12.3&29.0 &7.8&  -41.7\%&-26\% &-70.4\%\\
 \gemmatwo{} &  27.4&40.6 &35.2&  28.4&51.8 &36.4&  +3.7\%&+28\% &+3.3\%&  27.4&40.6 &35.2&  16.4&29.1&15.2&  \underline{20.5}&33.0 &20.3&  16.4&29.1 &15.2&  -40.2\%&-29\% &-56.9\%\\
 \qwen{} &  24.8&32.9 &29.8&  24.5&39.7 &30.0&  \underline{-1.3\%}&+\underline{21}\% &\textbf{+0.6\%}&  24.8&32.9 &29.8&  11.8&18.4 &9.5&  15.4&22.2 &12.5&  11.8&18.4 &9.6&  -52.6\%&-44\% &-68.2\%\\
 \jamba{} &  27.0&\underline{47.0} &38.5&  27.3&\textbf{67.2} &39.9&  \textbf{+1.0]\%}&+43\% &+3.5\%&  27.0&\underline{47.0} &38.5&  15.2&\underline{35.2} &\textbf{21.7}&  18.3&\textbf{38.3} &\textbf{26.0}&  15.2&\textbf{35.2} &\textbf{21.7}&  -43.9\%&-25\% &-43.6\%\\
 \commandr{}&  \underline{28.6}&\textbf{47.5} &\underline{40.4}&  \underline{29.7}&\underline{57.5} &\underline{40.7}&  +3.7\%&+\underline{21}\% &\underline{+0.7\%}&  \underline{28.6}&\textbf{47.5} &\underline{40.4}&  \underline{16.5}&33.1&\underline{18.5}&  20.1&\underline{37.5} &\underline{23.4}&  \underline{16.5}&\underline{33.1} &\underline{18.5}&  -42.3\%&-30\% &-54.2\%\\
 \gptfour{}&  \textbf{35.2}&\textbf{47.5} &\textbf{43.1}&  \textbf{36.3}&55.2 &\textbf{44.4}&  +3.0\%&+\textbf{16}\% &+2.9\%&  \textbf{35.2}&\textbf{47.5} &\textbf{43.1}&  \textbf{19.3}&\textbf{41.0}&13.3&  \textbf{22.7}&31.7 &17.0&  \textbf{19.3}&28.1 &13.3&  -45.1\%&-41\% &-69.2\%\\
 \bottomrule
 \end{tabular}
\caption{Results of the tests for changing the position of the time reference and the effect of shifting the referenced year. We report Contains, BEM, and OpenEval scores.}
\label{tab:position_error_all}
\end{sidewaystable}

Table~\ref{tab:order_all} contains the results for the event ordering task. 
\begin{sidewaystable}[ht]
\scriptsize
\centering
\begin{tabular}{l|ccc|ccc|ccc|ccc|ccc|ccc|ccc} 
\toprule
\textbf{Model} &  \multicolumn{21}{c}{\textbf{$\leftrightarrow$Event Ordering}}\\ \hline
               &  \multicolumn{3}{c|}{0}&  \multicolumn{3}{c|}{1}&  \multicolumn{3}{c|}{5}&  \multicolumn{3}{c|}{10}&  \multicolumn{3}{c|}{30}&  \multicolumn{3}{c|}{100}&  \multicolumn{3}{c}{Diff.{[}100,0{]}}\\ \hline
 &  Cont. & BEM&OE&  Cont. & BEM&OE&  Cont. & BEM&OE&  Cont. & BEM&OE&  Cont. & BEM&OE&  Cont. & BEM&OE& Cont. & BEM&OE\\ \hline

 \llamathreeone{} &  49.9&55.3 &3.1&  51.6&55.5 &2.8&  50.8&54.9 &3.0&  51.6&55.5 &2.8&  51.0&54.6 &3.3&  49.8&51.8 &3.4& \textbf{-0.2\%}&$\downarrow$\underline{6}\% &\textbf{-12.0\%}\\
 \gemmatwo{} &  36.3&49.5 &0.9&  49.4&57.5 &2.6&  36.8&46.9 &1.6&  49.4&57.5 &2.6&  42.1&50.9 &1.9&  49.4&57.5 &2.6&  -36.1\%&$\downarrow$16\% &-178.6\%\\
 \qwen{} &  48.4&46.9 &\underline{4.5}&  \textbf{81.1}&61.7 &\underline{5.3}&  \underline{57.5}&50.0 &\underline{4.8}&  \textbf{81.1}&61.7 &\underline{5.3}&  \underline{74.4}&59.3 &\underline{6.0}&  \underline{81.1}&61.7 &\underline{5.3}&  -67.6\%&$\downarrow$32\% &\underline{-17.8\%}\\
 \jamba{} &  49.0&48.5 &0.8&  49.5&59.6 &1.2&  50.2&49.7 &1.1&  49.5&49.6 &1.2&  51.5&50.7 &0.8&  49.5&49.6 &1.2&  \underline{-1.0\%}&$\downarrow$\textbf{2}\% &-50.0\%\\
 \commandr{}&  \underline{52.8}&\underline{66.6} &2.9&  50.6&\underline{65.0} &3.0&  54.4&\underline{67.3} &3.7&  57.8&\underline{70.1} &3.1&  64.1&\underline{73.2} &4.3&  70.4&\underline{77.3} &4.7& -33.5\%&$\downarrow$16\% &-61.4\%\\
 \gptfour{}&  \textbf{53.5}&\textbf{69.2} &\textbf{6.6}&  \underline{56.9}&\textbf{72.1} &\textbf{8.2}&  \textbf{64.0}&\textbf{75.2} &\textbf{7.0}&  \underline{70.1}&\textbf{78.3} &\textbf{7.1}&  \textbf{79.4}&\textbf{82.6} &\textbf{7.2}&  \textbf{86.6}&\textbf{86.7} &\textbf{9.2}& -61.8\%&$\downarrow$25\% &-39.2\%\\     
\bottomrule
\end{tabular}
\caption{Results for the event ordering tasks. 0,1,5,10,30,100 is the distance in years between the compared events.}
\label{tab:order_all}
\end{sidewaystable}

Table~\ref{tab:inverse_all} contains the results for the Temporal inverse tests. 
\begin{table*}[ht]
\scriptsize
\centering
\begin{tabular}{l|ccc|ccc|ccc}
\toprule
\textbf{Model} &\multicolumn{9}{c}{\textbf{$\leftrightarrow$Reversal}}\\ \hline
               &  \multicolumn{3}{c|}{Fwd}&  \multicolumn{3}{c|}{Fwd $\cap$ Inv}&  \multicolumn{3}{c|}{Diff.}\\ \hline
 &  Cont.& BEM&OE&  Cont.& BEM&OE& Cont.& BEM&OE\\ \hline

 \llamathreeone{} &  3.4&16.5 &7.3&  0.9&6.5 &3.0&  \textbf{-72.6\%}&-61\% &-55.6\%\\
 \gemmatwo{} &  7.6&25.1 &15.2&  1.6&9.5 &6.0&  -79.3\%&-62\% &-60.6\%\\
 \qwen{} &  5.6&15.6 &12.1&  1.2&4.5 &3.9&  -77.8\%&-71\% &-67.7\%\\
 \jamba{} &  8.8&\underline{35.5} &\underline{23.7}&  1.8&\underline{15.9} &9.0&  -79.9\%&-55\% &-62.1\%\\
 \commandr{}&  \underline{10.7}&29.8 &22.4&  \underline{2.3}&15.0 &\underline{10.2}& -78.6\%&-\underline{50}\% &\underline{-54.7\%}\\
 \gptfour{}&  \textbf{16.5}&\textbf{36.9} &\textbf{31.4}&  \textbf{4.5}&\textbf{20.4} &\textbf{16.5}& \underline{-72.9\%}&-\textbf{45}\% &\textbf{-47.4\%}\\
 \bottomrule
 \end{tabular}
\caption{Results of the tests for the temporal reversal test. We report Contains, BEM, and OpenEval scores.}
\label{tab:inverse_all}
\end{table*}

Table~\ref{tab:facts_all} contains the results for the fact checking and event dating tests. 
\begin{table*}[ht]
\scriptsize
\centering
\begin{tabular}{l|ccc|} 
\toprule
\textbf{Model} & \multicolumn{3}{c|}{\textbf{$\uparrow$Fact Checking}}\\ \hline
               & &&\\ \hline
 &  Cont. &BEM& OE\\ \hline

 \llamathreeone{} & 29.1 &73.1& 34.2\\
 \gemmatwo{} & 39.9 &74.7& 42.6\\
 \qwen{} & \textbf{74.7} &\underline{92.4}& 31.4\\
 \jamba{} & \underline{65.5} &\textbf{97.5}& \textbf{52.5}\\
 \commandr{}& 46.5 &75.4&\underline{45.7}\\
 \gptfour{}& 33.1 &65.3&36.5 \\     
\bottomrule
\end{tabular}
\caption{Results for the fact checking task. }
\label{tab:facts_all}
\vspace*{-2mm}
\end{table*}

\section{The Date-Match Metric}
\label{sec:appendix_datematch}
Although the human alignment results on the Event Dating task (Figure~\ref{tab:metrics_alignment}) suggest that Contains and OpenEval would be suitable choices, we manually found several cases where both OE and contains did not correctly identify the correct answers. This was usually the case when models did not adhere to the format specified in the prompt. Considering \gptfour{}'s answer: "The event you're referring to is the establishment of Reykjavík, which occurred on 18 August 1786 [...]". Given the ground-truth string of "18-08-1786", both OE and contains did not correctly identify the answer as correct. The contains metric did understandably fail as soon as the answer's format did not precisely match the ground-truth format applied for the event dating task ("dd-mm-yyyy"). 
We, therefore, use our own date-matching metric that we built using the python dateutil\footnote{\url{https://github.com/dateutil/dateutil/}} library, which tries to parse a date object from the predicted answers. If the ground-truth date and the parsed date match, the answer is scored with a 1 and otherwise with a 0. After manually inspecting 100 predicted answers for each model, we find ~20 different ways to write the dates being used and verify that our metric correctly parses all of them (Table~\ref{tab:date_match_exmaples}). Additionally, we handle incomplete predictions: Per default, the dateutil package that we use parses “April 2020” to “1st of April 2020”, so in case of a missing day, it will use the first day of the month. For ambiguous predictions such as 11-10-2020 we assume the format provided in the prompt. In our prompt for the event dating task, we specify the format (“dd-mm-yyyy”). So 11-10-2020 would be parsed as the 11th of October.

\begin{table*}[ht]
\centering
\small
\begin{tabular}{l} \toprule
\textbf{Date Format}                                                                            \\ \hline
"this is a full sentence July 18, 1956."                                                        \\
"October 19, 1763."                                                                             \\
"March 24, 1935"                                                                                \\
"December 2019"                                                                                 \\
"2019"                                                                                          \\
"Dec, 2019"                                                                                     \\
"September 27, 1941"                                                                            \\
"October 17, 1961 (10-10-1961)"                                                                 \\
"27 February 1977"                                                                              \\
"24 May 1899."                                                                                  \\
"2nd December 1959."                                                                            \\
"10th of July, 1806."                                                                           \\
"16th of October, 1756"                                                                         \\
"12-December-1957."                                                                             \\
"01-01-1930."                                                                                   \\
"01-09-1950."                                                                                   \\
"14-11-1972."                                                                                   \\
"100712"                                                                                        \\
"01012022"                                                                                      \\
"28/3/1941"                                                                                     \\
"04 03 1809"                                                                                    \\
"20111104"                                                                                      \\
"2011-11-04"                                                                                    \\
"1502-02-11"                                                                                    \\
"Jan 9, 2021"                                                                                   \\
"Jun 11, 2023"                                                                                  \\
"9 Jan 2021"                                                                                    \\
"21 NOV 1859"                                                                                   \\
"9-Jan-2021"                                                                                    \\
"11-Jun-2023"                                                                                   \\
"The event occurred on 23-25-2020 (DD-MM-YYYY)."                                                \\
"The Peking Opera was born on 1759-01-01, which is January 1, 1759"                             \\
"The Bhadla Solar Park was commissioned on March 25, 2012. Therefore, the event happened on 25" \\ \bottomrule
\end{tabular}
\caption{Different formats of dates given by the models in this study. We made sure that the date-match metric properly parses at least these variations.}
\label{tab:date_match_exmaples}
\end{table*}

\section{Prompts}
Prompts for the tasks are shown in Table~\ref{tab:prompts}. Additionally, we include the utilized system prompts in Table~\ref{tab:prompt_system}.

\begin{table*}[ht]
    \centering
    \renewcommand{\arraystretch}{1.2}
    \begin{tabularx}{\linewidth}{l|X}
        \toprule
        \textbf{Task} & \textbf{Prompt Template} \\
        \hline
        QA & Please answer the question:\newline \{question\} \newline Answer: \\
        \hline
        Event Ordering & Please answer the question with 'True' or 'False'.\newline Question: Did A happen before B? \newline \newline A: \{event1\} \newline B: \{event2\} \newline \newline Answer: \\
        \hline
        Fact Checking & Please answer the claim with 'True', 'False' or 'Conflicting'.\newline Claim: \{claim\} \newline Answer: \\
        \hline
        Event Dating (Day) & Here is an event:\newline \{event\} \newline Please answer with the date on which the event happened (DD-MM-YYYY). \newline Answer: \\
        \hline
        Event Dating (Month) & Here is an event:\newline \{event\} \newline Please answer with the date on which the event happened (MM-YYYY). \newline Answer: \\
        \hline
        Event Dating (Year) & Here is an event:\newline \{event\} \newline Please answer with the date on which the event happened (YYYY). \newline Answer: \\
        \hline
        Completion & Please complete the following sentence:\newline \{question\} \\
        \bottomrule
    \end{tabularx}
    \caption{Prompt Templates for Different Tasks.}
    \label{tab:prompts}
\end{table*}

\begin{table*}[ht]
    \centering
    \renewcommand{\arraystretch}{1.2}
    \begin{tabularx}{\linewidth}{l|X}
        \toprule
        \textbf{Mode} & \textbf{Description} \\
        \hline
        Default & You are a helpful assistant. \\
        \hline
        Historian & Provide direct and concise answers to historical or temporal questions. You are a historian specializing in temporal question answering. Please avoid speculation and present verified historical knowledge wherever possible. \\
        \hline
        CoT & You are a helpful assistant that thinks step by step. \\
        \bottomrule
    \end{tabularx}
    \caption{Different system prompts.}
    \label{tab:prompt_system}
\end{table*}

\end{document}